\documentclass{article}
\usepackage{ijcai25}

\usepackage{times}
\usepackage{soul}
\usepackage{url}
\usepackage[hidelinks]{hyperref}
\usepackage[utf8]{inputenc}
\usepackage[small]{caption}
\usepackage{graphicx}
\usepackage{amsmath}
\usepackage{amsthm}
\usepackage{booktabs}
\usepackage{algorithm}
\usepackage{algorithmic}
\usepackage[switch]{lineno}
\usepackage{multirow}
\usepackage{makecell}
\usepackage{amssymb}
\usepackage{tikz}
\usetikzlibrary{shapes,arrows,positioning,backgrounds}

\title{Playing games with Large language models: Randomness and strategy}

\author{
Alicia Vidler$^1$
\and
Toby Walsh$^2$\and
\affiliations
$^1$UNSW\\
$^2$UNSW\\
\emails
\{a.vidler, t.walsh\}@unsw.edu.au,
}

\begin{document}

\maketitle

\begin{abstract}
Playing games has a long history of describing intricate interactions in simplified forms. In this paper we explore if large language models (LLMs) can play games, investigating their capabilities for randomisation and strategic adaptation through both simultaneous and sequential game interactions. We focus on GPT-4o-Mini-2024-08-17 and test two games between LLMs: Rock Paper Scissors (RPS) and  games of strategy (Prisoners Dilemma PD).  LLMs are often described as stochastic parrots \cite{GebruParrot}, and while they may indeed be parrots, our results suggest that they are not very stochastic in the sense that their outputs - when prompted to be random - are often very biased.  Our research reveals that LLMs appear to develop loss aversion strategies in repeated games, with RPS converging to stalemate conditions while PD shows systematic shifts between cooperative and competitive outcomes based on prompt design.   We detail programmatic tools for independent agent interactions and the Agentic AI challenges faced in implementation.  We show that LLMs can indeed play games, just not very well. These results have implications for the use of LLMs in multi-agent LLM systems and showcase limitations in current approaches to model output for strategic decision-making.

\end{abstract}

\section{Introduction}
Large language models have seemingly become ubiquitous since first garnering user attention in 2022, and have gone on to established themselves as the "most influential and widely-adopted AI technology to date" \cite{ShakiCogEff}.  Game theory is a long standing, rich area of scientific and economic study, where the use of stylised human participants suggest they may benefit from LLM incorporation.  Indeed LLMs have already proved themselves useful in social science experiments and societal simulation \cite{Guo2024}. 

However there remains questions that, they may be acting as a "stochastic parrot" \cite{GebruParrot} rather than replicating human decisions with fidelity.  Large language models have been shown to be suboptimal at producing random sequences (\cite{Liu2024}, \cite{Renda2023}, \cite{Gu2024}and [annon, forthcoming]).  Work by [annon, forthcoming] show that OpenAI LLM models have significant variations in distributional outcomes depending on the model, and the model subversion.  Also, results are dependent on the method used to query an API (i.e. independent sampling of an API or batch sampling produce different results).   In game theory, an important game outcome is that of the "mixed strategy" - whereby a player randomly chooses between different pure strategies according to specific probabilities.  Fundamental to stage games such as RPS, randomness thwarts the problem that being predictable players leads to losing. Mixed strategies in games are also often optimal in zero-sum games to prevent exploitation. 

This raises two important questions - given the concerns around LLM ability to replicate random variables, can LLMs learn optimal mixed strategies without true randomness? Can they develop meta-strategies that compensate for their randomness limitations?  In this paper we investigate whether an LLM, acting as an independent agent, can play a series of increasingly strategic and complex games.  In answering these questions, the ability of LLMs to effectively play games with each other can indicate several important capabilities that are valuable beyond just game play.  Emergent properties of agent co-ordination or loss aversion, for example, would be of use to agent based modellers utilising LLMs. 

Our contribution to these fields is to explore if LLMs can successfully be deployed to play multi-agent games  using two common game theoretic stage games: RPS and strategic games like PD.  We test both one-shot and repeated versions of these games, with RPS requiring uniform distribution strategies and PD featuring a dominant strategy alongside a Pareto optimal outcome.  We explore a variety of prompts, and find, as expected, LLMs remain susceptible to prompts, but not always in ways expected.  Games are constructed using LangChain as an agentic method and we make use of State of the Art GPT-4o-Mini-2024-07-18 for all tests, given previously reported poor results for other models to produce random sequences and the relevance of model subversion testing \cite{vidler2025evaluatingbinarydecisionbiases}.



\section{Background and Recent Literature}

\subsection{Game theory}
Games of strategy cover many forms with early work pioneered by \cite{vonNeumannMorgenstern+2004} (single shot, pure and mixed strategies) and numerous extensions including repeated games \cite{AumannRepeated}. A central idea in game theory is the notion of equilibrium \cite{nash1951non}, an outcome where neither player can unilaterally deviate and improve their payoff. Embedded in this concept is the idea that each player's payoff (or return for playing the game) has some dependence on another player's response. Probabilistic notions of Bayesian games was explored by \cite{Harsanyi} where he extended the ideas and importance that probability or randomness can play in a game's equilibrium. Introducing these notions to agent based systems \cite{Wooldridge_2009} also details associated concepts of Pareto optimality (where it is not possible to make any one individual better off without making at least one other individual worse off) and agent coalition formation, especially as they relate to computers playing games (\cite{Shoham_2008}). Often, however, strategic games have no clear consistent “best strategy” (or rather, no pure equilibrium). Instead, games often times have a mixed strategy as a form of best play, focusing on the importance of probability and randomness in game playing.  We consider if indeed LLM agents can learn from repeated games, do they possibly exhibit some form of co-ordination (in the vein of \cite{harsanyi1988general} or loss aversion in the spirit of \cite{axelrod1984evolution}. We consider two stage games here: RPS (relying on \cite{Zhou_2015}) and PD games (\cite{FRANK1993247} and \cite{BROSIG2002275} amongst others). 

\subsection{Large Language Models}

Large Language Models (LLMs) represent a significant advancement in natural language processing, built upon transformer architectures that utilise self-attention mechanisms to process sequential data (see \cite{vaswani2023attentionneed} for more details).  Pre-trained Large Language Models (LLMs), such as GPT-3 \cite{bubeck2023sparks} and GPT 4 \cite{Luo2023}, have demonstrated proficiency in tasks like dialogue generation and natural language interactions. LLMs excel in generating coherent text across various domains (\cite{chang2023survey}, yet they face notable limitations in areas such as reasoning, numerical understanding, and optimal prompting methods \cite{srivastava2023imitationgamequantifyingextrapolating}, \cite{zhang2024llmmastermindsurveystrategic}.  Specifically, 'reasoning' in negotiation remains a challenge \cite{Abdelnabi}, \cite{zhang2024llmmastermindsurveystrategic} - show LLMs under performing on negotiation tasks.  However, work by \cite{Faria2022} shows that LLMs can adeptly learn to interpret corporate financial reports.  Working with LLMs via \textbf{prompt design} remains an evolving research area. Studies such as \cite{Zamfirescu-Pereira2023} have observed that non-expert users often adopt opportunistic, rather than systematic, approaches to prompting.  One area that garners a lot of curiosity is the phenomena of "hallucinations".  Several approaches and interpretations exist (\cite{zhang2022},\cite{Wei2022a}), and suggestions around faithful reasoning are posited \cite{creswell2022}.  Work looking at the improvement from multi step reasoning are covered in \cite{Fu2023}.  

\subsection{Computational Randomness}
Research on LLM randomness has tended to focus on the inability of specific generic LLM models (GPT-3.5, LLAMA 7B etc) to produce random numbers (\cite{Renda2023}, \cite{Harrison}, \cite{Liu2024}), generate probabilistic text \cite{Tjuatja2024}, \cite{Renda2023}, \cite{Peeperkorn2024} or more broadly, replicate behavioural simulations \cite{Gu2024}. Whilst such work often fails to identify the precise versions of the LLMs tested, results suggest that LLMs perform poorly at sampling from a specific distribution (e.g. normal, uniform etc), LLMs cannot reason well \cite{Imani2023} nor reproduce a requested distribution \cite{Renda2023}. Recent work by [annon, forthcoming] has shown that LLM models of the OpenAI family give very different results in sampling random variables (even between model subversions of GPT4) but that most models showed significant bias and don't reliably produce uniform distributions or maintain Markovian properties, despite tuning for Temperature parameters. Sampling methods (one-shot or few-shot sampling) significantly affected model outputs [annon, forthcoming].

Recent work found that LLMs might even produce results with greater levels of bias than humans \cite{Tjuatja2024}, in contrast to \cite{Harrison}, who found that GPT-3.5 Turbo-0125 could under certain scenarios produce more "random" number generation than human subjects. Studies show LLMs struggle with sampling from specific distributions \cite{Imani2023}, \cite{Renda2023}. \cite{GuoABMLLMijcai} provides a comprehensive survey detailing LLM agent research focusing on two areas broadly - agent decision making and that of agent simulation engines to replicate complex real-world environments, finding LLMs capable of "exceptional role-playing abilities."

Our contribution to this literature in this paper is to explore the use of LLMs in game playing simulations, focusing on the most successful of model at random sampling identified in [annon, forthcoming], namely GPT- 4o-Mini-2024-07-18.

\subsection{LLM Cognitive effects and LLM collaboration}
As LLM usage increases, so too are questions being asked by  researchers such as about their limitations and about theory of mind. Work delving into the cognitive effects of LLMs focused on earlier OpenAI models (GPT-3) in \cite{ShakiCogEff} found that LLMs were indeed prone to a number of "human cognitive effects." While \cite{AmosHumanLike} reports LLMs' fondness for the number 7 and other examples demonstrating that "ChatGPT's responses in several social experiments are much closer to those of humans than to those of fully rational agents," work by \cite{li-etal-2023-theory} utilised GPT-3.5-turbo-0301 and GPT-4-0314 and found evidence of emergent collaborative behaviours between agents when given specified collaborative tasks.

\subsection{Agentic LLMs}
The integration of LLMs within agent-based frameworks enables practical implementation of game-theoretic interactions between independent LLM instances. One such method is Concordia \cite{vezhnevets2023generative}, which has been shown to create Generative Agent-Based Models (GABMs).  Agents show abilities to "apply common sense" and "act reasonably" within simulated environments \cite{park2023generative}. Recent studies \cite{Abdelnabi}, \cite{zhang2024llmmastermindsurveystrategic} reveal that current LLMs under-perform in negotiation tasks and strategic reasoning within agent-based systems. Work such as \cite{WuAutoGen} has shown the possibility of generic agents using LLMs.  Beyond Concordia, several commercial frameworks exist for managing meta-agents, with LangChain \cite{Langchain} being a prominent Python framework for LLM application development \cite{LangChain2023creating}.  In this work we make use of LangChain.

\section{Method: Games}

\begin{figure}[t]
        \centering
        \resizebox{\columnwidth}{!}{
        \begin{tikzpicture}[
            node distance=1.5cm,
            box/.style={rectangle, draw, minimum width=1.2cm, minimum height=0.7cm},
            llm/.style={box, fill=yellow!30},
            player/.style={box},
            evaluate/.style={box},
            history/.style={box},
            thick arrow/.style={->, very thick, color=pink!60!red},
            bidir/.style={<->, bend left=15}  
        ]
        
        \node[llm] (llm) {LLM};
        \node[player, right=of llm, yshift=1.2cm] (p1) {P1};
        \node[player, right=of llm] (p2) {P2};
        \node[player, right=of llm, yshift=-1.2cm] (p3) {P3};
        \node[evaluate, right=of p2] (eval) {Eval};
        \node[history, right=of eval] (hist) {Hist};
        
        \foreach \y/\p in {20/p1, 0/p2, -20/p3} {
            \draw[bidir] (llm) to[bend left=\y] (\p);
        }
        
        \draw[->] (p1) -- (eval);
        \draw[->] (p2) -- (eval);
        \draw[->] (p3) -- (eval);
        \draw[->] (eval) -- (hist);
        \draw[thick arrow] (hist) to[bend right=45] (llm);
        
        \end{tikzpicture}
        }
        \caption{Game flow diagram showing bidirectional, independent, LLM interactions with three players, evaluation, and history feedback loop (applicable only for \textit{Repeated games}.}
        \label{fig:game-flow}
    \end{figure}
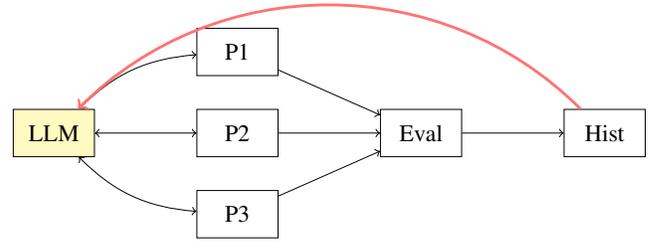

We developed our game-playing framework using LangChain and GPT-4o-Mini-2024-08-17, implementing a standardised model flow for all games as shown in Figure \ref{fig:game-flow}. We conduct 100 simulations for both one-shot and repeated games, providing adequate statistical power to analyse distributional outcomes and strategic patterns. Thus, our findings reveal that deviations from anticipated distributions achieve statistical significance at or beyond the 95\% confidence level.


The framework consists of two core components: a Player module containing game-specific prompts, and an Evaluation module managing game payoffs and evaluation logic. For single-shot games, the framework executes each simulation independently, while repeated games maintain a historical record that feeds into subsequent LLM interactions. Each cycle completion represents one complete game simulation.


\textbf{OpenAI LLM Caching} \\
OpenAI's API employs key-specific caching to optimise response times for identical prompts. While this caching cannot be disabled programmatically, we implemented three strategies to ensure independent sampling in our LLM experiments. First, we generated new API keys regularly to avoid retrieving cached results. Second, we appended unique identifiers to each prompt for one-shot games. And, thirdly, we appended complete game histories to prompts for repeated games.  These measures effectively prevented cache retrieval, though they increased processing time and API costs. The impact of caching on prior research into OpenAI's random number generation capabilities remains unclear and warrants further investigation.

\subsection{Game rules: Prompts}
Following in the vein of \cite{Zamfirescu-Pereira2023}, we test a variety of prompts for each game opportunistically. Prompts must describe the game structure (\textit{e.g. "You are playing the strategic game called Rock Paper Scissors and you need to choose what your play will be."}, the payoff function and the desired model return value (e.g. choose Rock or Paper etc).  Some research has shown that LLMs can be sensitive to the syntax of prompts, particularly as regards random variable generation \cite{Liu2024}. In developing our method here we tested a number of prompts opportunistically to check for any large observed variations. We have provided the prompts for each test below.

\subsection{Rock Paper Scissors:  Games of Uniform distribution}
RPS is a simple game where each player can choose between three options, with a simple hierarchy of win vs loss.  It is a stage game for non-cooperative strategic interactions. 
 The rules define a zero-sum game with a unique mixed strategy Nash equilibrium where each player uniformly randomises over all actions, playing each with probability \(\frac{1}{3}\) \cite{Zhou_2015}.  Formally the game, \(G(RPS)\), is defined as let \(G = (N, A, u)\) be a symmetric two-player normal-form game where: \(N = \{1,2\}\) is the set of players, \(A = \{R,P,S\}\) is the action set for each player, representing Rock, Paper, and Scissors respectively and \(u: A \times A \rightarrow \mathbb{R}^2\) is the payoff function which can be expressed as: 
$$
\begin{array}{c|ccc}
 & R & P & S \\
\hline
R & (0,0) & (0,1) & (1,0) \\
P & (1,0) & (0,0) & (0,1) \\
S & (0,1) & (1,0) & (0,0)
\end{array}
$$

\begin{quotation}
    
\subsubsection*{Base Prompt (1)}
    \textit{You are playing Rock Paper Scissors. Rules: (1) Choose: 'R' (Rock), 'P' (Paper), or 'S' (Scissors) (2) Winning conditions: Rock crushes Scissors, Paper covers Rock, Scissors cuts Paper (3) Analyze the game history to identify patterns (4) Only respond with a single letter: 'R', 'P', or 'S', no explanation}
     
\subsubsection*{Prompt (2) - re ordering of options}
    \textbf{Rock first}: " \textit{You are playing \textbf{Rock} Paper Scissors where winning awards 1 point and losing gives 0 points. Paper beats Rock by covering it, Rock beats Scissors by breaking them, and Scissors beats Paper by cutting it. Matching moves result in a tie where both players get 0 points. The payoffs are written as (Player 1 score, Player 2 score) for each possible combination. Only respond with a single letter: R, P, S} "

    \textbf{Paper first}: The same prompt as above but with Paper first e.g "\textbf{Paper} Scissors Rock" and "Choose one of: \textbf{'P' (Paper)} or 'S' (Scissors)" , "Only respond with a single letter: \textbf{P}, S, or R "
        
    \textbf{Scissors first}: Much like Paper above, "\textbf{Scissors} Rock Paper." and "Choose one of: \textbf{'S' (Scissors)} or R' (Rock) or 'P' (Paper)." , "Only respond with a single letter: \textbf{S}, R, or P "

\subsubsection*{Prompt (3a): Classic design reworded}
        "\textit{You are playing Rock Paper Scissors. Make strategic choices based on game patterns and theory. Rules: Choose one of: 'P' (Paper), 'R' (Rock), or 'S' (Scissors). Payoff, Paper beats Rock, Rock beats Scissors, Scissors beats Paper, all other combinations are a tie. Only respond with a single letter: P, R, or S}"
    
\subsubsection*{Prompt (3b): Added "random" to prompt}
        "\textit{You are playing Rock Paper Scissors. Make strategic choices based on game patterns and theory. Rules: \textbf{Randomly} choose one of: 'P' (Paper), 'R' (Rock), or 'S' (Scissors). Payoff, Paper beats Rock, Rock beats Scissors, Scissors beats Paper, all other combinations are a tie. Only respond with a single letter: P, R, or S}"
        
\subsubsection*{Prompt (3c): Added "random" and optimal Strategy recommendation}
        "\textit{You are playing Rock Paper Scissors. Make strategic choices based on game patterns and theory. Rules: \textbf{Randomly} choose one of: 'P' (Paper), 'R' (Rock), or 'S' (Scissors). Payoff, Paper beats Rock, Rock beats Scissors, Scissors beats Paper, all other combinations are a tie. \textbf{The optimal strategy is to randomise your selection of R,P,S}. Only respond with a single letter: P, R, or S}"   

\end{quotation}

\section{Results}

\subsection{RPS: One-Shot RPS games with 2 virtual players - Non uniform play}
In Table \ref{tab:1s_rps_strategies} we see that regardless of prompts, none of the tests can come close to a uniform distribution, which is of course the optimal strategy result. Given a sample of 200 (one observation per player per 100 games), proportions falling between 25.4\% and 41.3\% cannot be considered statistically significantly different from the expected uniform distribution rate of \(\frac{1}{3}\) at \(p < 0.05\). 


\begin{table}[ht]
    \centering
        \begin{tabular}{lccccr}
        \small
        Temperature = 1 \\
        
        100 - One-Shot Simulations \\
        \hline
        Prompt & R & P & S & TIE *\\ 
        \hline
        P1 Base case & 79\% & 13\% & 9\% & 68\% \\ 
        \hline
        P2 Classic design & 61\% & 25\% & 14\% & 46\% \\ 
        P2 Syntax order changed \\ (Paper first) & 71\% & 13\% & 17\% & 54\% \\ 
        P2 Syntax order changed \\ (Scissors first) & 80\% & 14\% & 7\% & 66\% \\ 
        \hline 
        P3a Classic & 75\% & 10\% & 16\% & 55\% \\ 
        P3b "Random" & 90\% & 7\% & 4\% & 79\% \\ 
        P3c  Strategy Instruction & 87\% & 4\% & 10\% & 78\% \\ 
        \hline
        Average & 77\% & 12\% & 11\% & 64\% \\
        Player 1 Avg Score: 17.2 \\
        Player 2 Avg Score: 19.5
       
        \end{tabular}
    \caption{One-shot RPS Distribution as a \% of game plays and and *Tie Rates as a \% of overall numbers of games}
    \label{tab:1s_rps_strategies}
\end{table}

\subsubsection{P1: Base case}
Between P1 and P2 (Classic design) and P3a (Classic design), we see some variation in results from what should be essentially the same test, though the order of plays is similar (Rock dominates) - none produce a uniform distribution.  Further more, in a classic game of RPS we would expect to see \( \frac{1}{3}\) of results be ties or draws.  Here we see numbers ranging from 46\% to 79\% draws.  We also observed no meaningfully changes in distributions for larger samples (e.g P2 Classic was simulated for 1,000 independent samples of the game with RPS results of (69\%, 17\% and 14\% respectively, with 52\% being ties). All observed proportions lie outside the 25.4\% - 41.3\% range of statistical similarity to the expected uniform distribution \( (p < 0.05) \), indicating systematic deviation from random play.

\subsubsection{Syntax}
Given the dominance of "Rock", P2 also shifts the prompts to put "Paper" first and then "Scissors". In both cases, changing the order decreases the selection of the first variable, though distributions remain non-uniform. This is somewhat counterintuitive but does suggest order of variables is not responsible, entirely, for "Rocks" dominance. 

\subsubsection{Prompts with Strategic Hints}
Our final test phase (P3a to c) tested whether explicit randomisation cues would improve play distribution. We introduced two variations: adding the term "random" to the instructions and providing information about optimal strategy. Surprisingly, neither modification led to more balanced play distributions. The models continued to show strong preferences for certain moves, suggesting that explicit hints about randomisation and optimal strategy do not effectively influence the model's decision-making process.

\subsubsection{1-shot RPS Conclusion}
We show it is technically feasible to construct one-shot games between two LLMs, their performance on RPS reveals significant limitations in handling uniform distribution games. Although prompt syntax had some influence on play patterns, none of our prompt variations achieved the game-theoretic optimal solution of uniform distribution (1/3 probability for each move).  Most notably, the LLM appeared resistant to explicit strategic guidance, maintaining biased play patterns even when prompted with optimal strategy suggestions. This suggests fundamental limitations in the LLMs' ability to implement probabilistic decision-making, in line with literature.  In the next section we will consider repeated games of RPS.

\subsection{Repeated RPS Games: Learning and Stalemate Emergence}

\textbf{Method}\\
    The tests detailed in Table \ref{tab:1s_rps_strategies} are repeated for simulations of 100 repeated games and reported in Table \ref{tab:100s_rps_strategies}. In these extended games, each agent receives a complete history of both players' moves and payoffs up to the current simulation (e.g., \('round': 1, 'moves': 'Player_1': 'R', 'Player_2': 'R', 'payoffs': 'Player_1': 0, 'Player_2': 0 \)). To avoid any potential response caching by the GPT-4o-Mini-2024-08-17 API, we implemented three controls mentioned earlier in the Methods section.

\subsubsection{Repeated RPS: 100 games Results}
Results in Table \ref{tab:100s_rps_strategies} continue to show persistent deviations from optimal strategy across all prompt variations.  The baseline prompt (P1) appears to show some balance with Rock and Paper selections (39\% and 38\% respectively) falling within statistical similarity to the expected uniform distribution (25.4\% - 41.3\%, \(p < 0.05\)), however, the Scissors selection (24\%) remains significantly below expected rates. While modifying syntax produces some statistically uniform selections, particularly in Rock and Paper choices for P2 Classic design (37\% and 41\%), each prompt variant maintains at least one significantly deviant selection rate. Attempts to influence randomisation through explicit instruction prove counterproductive, with the addition of the word "random" producing a significant Rock bias of 57\%. Likewise, direct optimal strategy guidance fails to achieve uniform distribution, showing significant deviations in both Rock and Scissors selections (49\% and 18\% respectively). The model's overall bias is evident in the average distributions, where only Rock (40\%) maintains statistical similarity to uniform distribution, while Paper and Scissors selections (43\% and 17\%) show significant deviation. These patterns of non-random selection are further emphasised by tie rates (51\%) significantly exceeding the expected one-third for uniform play. Values falling within ranges are represented in {\color{green}Green} in the table below.

       

\begin{table}[h]
    \centering
        \begin{tabular}{lccccr}
        \small
        Temperature = 1 \\
        100 Repeated Games \\
        \hline
        Prompt & R & P & S & TIE* \\ 
        \hline
        P1 Base case & {\color{green}39\%} & {\color{green}38\%} & 24\% & 47\% \\ 
        \hline
        P2 Classic design & {\color{green}37\%} & {\color{green}41\%} & 22\% & 54\% \\ 
        P2 Syntax order changed \\ (Paper first) & 47\% & {\color{green}29\%} & 25\% & 43\% \\ 
        P2 Syntax order changed \\ (Scissors first) & 19\% & 76\% & 5\% & 66\% \\ 
        \hline 
        P3 Classic design reworded & {\color{green}35\%} & 48\% & 18\% & 43\% \\ 
        P3 Add word "random" & 57\% & {\color{green}36\%} & 7\% & 49\% \\ 
        P3 Add Optimal \\ Strategy Instruction & 49\% & {\color{green}34\%} & 18\% & 43\% \\ 
        \hline
        Average & {\color{green}40\%} & 43\% & 17\% & 51\% \\
        Player 1 Avg Score: 25.3 \\
        Player 2 Avg Score: 26.5
       
        \end{tabular}
    \caption{Repeated (100) RPS Game Distribution as a \% of game plays and *Tie Rates as a \% of overall numbers of games}
    \label{tab:100s_rps_strategies}
\end{table}

\subsubsection*{Comparisons One-shot and Repeated games}
The behavioural patterns differ between repeated and single-shot games. In repeated play, the baseline prompt (P1) achieves balanced distribution between Rock (39\%) and Paper (38\%), with Scissors at 24\% - a clear shift from single-shot's Rock dominance (79\%). While syntax ordering between repeated games of P2 does not appear to have a large impact, the overall difference between results for one-shot and repeated versions of the game for Prompt 2 is notable: in one-shot format when "Scissors" appears first in the prompt the distribution was (R:80\%, P:14\%, S:7\%) as compared to the repeated game results of (R:19\%, P:76\%, S:5\%).  Whilst the distribution to Scissors is largely unchanged between game styles, the balance between Rock and Paper is reverse.  In the P3 group of test, we isolate explicit randomisation and see this reduces Rock selection from 90\% in single-shot to 57\% in repeated play. The average distribution metrics show improved balance in repeated games (R:40\%, P:43\%, S:17\%) versus single-shot (R:77\%, P:12\%, S:11\%), with tie rates moving toward the theoretical \(\frac{1}{3}\) (51\% versus 64\%). Average player scores also increase from below 20 in one-shot games to 25-26 points in repeated settings. Whilst still low, and below the game theoretic expected payoff, there is a relatively large improvement towards expected values.  To examine if these improvements strengthen over time, we expanded our analysis to 1,000-turn games.

\subsection{Emergence of Stalemate in 1000 turn repeated games}

The 1,000 turn repeated game setup is identical to that described for the 100 repeated games. New API keys are used to avoid any possible cache issues. We test for two features: (a) does the behaviour over the first 100 repeated games persist in the remaining 900 (stationarity) and (b) - given the relatively high tie rates, does this persist uniformly (i.e how many games does it take for player scores to double across the 1000 simulations). For ease of comparison, three prompts were tested at this level; P2 classic design , P3(c) Express Optimal Strategy Direction and an additional new prompt, P4, documented below. 


\textit{Note: Statistical tests are carried out for 100 repeated games (200 observations), 101-1000 games (1800 observations, range 30.7\% - 36.0\% at \(p < 0.05\)) and 1000 games (2000 observations, range 30.8\% - 35.9\% at \(p < 0.05\)). Values falling within ranges are represented in {\color{green}Green} in the table below.}

\begin{quotation}
    \subsubsection*{Prompt P4: Clear Points}
        "\textit{You are playing the strategic game called Rock Paper Scissors and you need to choose what your play will be. You can choose one choice from the following list: Rock, Paper or Scissor. Your payoff will depend on the other players choice too: Paper beats Rock and wins \textbf{1 point}, Scissors beats Paper and wins \textbf{1 point}, Rock beats Scissors and wins \textbf{one point}, all other combinations, and \textbf{a tie, win 0}. Only respond with a single letter: R, P, S}"
\end{quotation}


\subsubsection{How stationery are Game results}
The P2 Classic Design demonstrates evolving behaviour across extended game play.  Directly comparing 100 repeated games for P2 against the \textbf{first 100 repeated games out of 1000} for P2 shows an initial 100-game simulations yielded distributions of R:37\%, P:41\%, S:22\% with 54\% ties (Rock and Paper selections falling within statistical similarity to the expected uniform distribution again at \(p < 0.05\)). Table \ref{tab:1000_p2_classic_distribution} shows consistent results in games 1-100 as would be hoped.  However, comparing games 101-1000 to 1- 100 (in either group) reveal shifts in play patterns. Only the results for Rock fall within the statistical similarity range for the given sample size. 
These later games show reduced ties but increased concentration in Rock and Paper selections at Scissors' expense. The P3 Express Optimal Strategy Direction prompt is reported in Table \ref{tab:1000_p3_optimal_distribution}, with only Rock results (33\%) for 1-100 games falling within statistical similarity to the expected uniform distribution again at \(p < 0.05\)).  Prompt 4 shows no results within statistically similar ranges to the expected value of \(\frac{1}{3}\), for any sample size.

 


    \begin{table}[htpb]
        \raggedright
        \small
        \begin{tabular}{lcccccc}
        \hline
        P2 Classic & R & P & S & Tie**& P1* & P2* \\ 
        \hline
        Games 1-100 & {\color{green}41\%} & {\color{green}37\%} & 23\% & 46\% & 24 & 30 \\
        Games 101-1000 & {\color{green}35\%} & 53\% & 12\% & 46\% & n/a & n/a \\
        \hline
        Total & 36\% & 51\% & 13\% & 45.7\% & 275 & 268  \\
        \hline
        \end{tabular}
        \caption{1000 Game Distribution of Moves in P2 Classic Design: Early vs Later Games. Note (*) are player scores, (**) Tie Rates as a \% of overall numbers of games}
        \label{tab:1000_p2_classic_distribution}
    \end{table}

\begin{table}[htpb]
    \raggedright
    \small
    \begin{tabular}{lcccccc}
    \hline
    P3c  & R & P & S & Tie**& P1* & P2* \\ 
    \hline
    Games 1-100 & 47\% & {\color{green}33\%} & 20\% & 48\% & 26 & 26 \\
    Games 101-1000 & 39\% & 56\% & 6\% & 50\% & n/a & n/a \\
    \hline
    Total & 39\% & 53\% & 7\% & 50\% & 248 & 256  \\
    \hline
    \end{tabular}
    \caption{1000 Game Distribution of Moves in P3 with Optimal Strategy Instruction: Early vs Later Games. Note (*) are player scores, (**) Tie Rates as a \% of overall numbers of games}
    \label{tab:1000_p3_optimal_distribution}
\end{table}

   \begin{table}[htpb]
       \centering
           \begin{tabular}{lcccccc}
           \hline
           P4 & R & P & S & Tie & P1* & P2* \\ 
           \hline
           Games 1-100 & 77\% & 22\% & 2\% & 72\% & 13 & 15\\
           Games 101-1000 & 89\% & 11\% & 1\% & 81\% \\
           \hline
           Total & 87\% & 12\% & 1\% & 80\% & 92 & 104\\
           \hline
           \end{tabular}
       \caption{1000 Game Distribution of Moves in P4 with Clear Points Structure: Early vs Later Games. Note (*) are player scores}
       \label{tab:1000_p4_points_distribution}
    \end{table}

\begin{table}[htpb]
   \centering
       \begin{tabular}{lcccc}
       \hline
       & \multicolumn{2}{c}{Player1} & \multicolumn{2}{c}{Player2} \\
       \cline{2-5}
       Wins & Game No. & increment & Game No. & increment\\
       \hline
       20 & 123 & 123 & 113 & 113 \\
       40 & 229 & 106 & 206 & 93 \\
       60 & 464 & 235 & 352 & 146 \\
       80 & 740 & 276 & 593 & 241 \\
       100 & n/a & n/a & 930 & 337 \\
       \hline
       \end{tabular}
   \caption{P4 1000 turn Achievement of Win Milestones by Players: Game Number and Duration}
   \label{tab:1000_p4_player_milestones}
\end{table}

\subsubsection*{Results: Loving Stalemates or Learning to avoid losing?}
Results from extended 1000-game simulations (P2, P3c, P4) demonstrate patterns of strategic stagnation, as simulations progress, tie rates increase due to agents repeatedly selecting identical moves despite their history of zero-point outcomes. P4's distinctive payoff structure (1 for wins, 0 for ties or losses) illustrates this evolution clearly, with tie rates rising from 72\% to 81\% (Table \ref{tab:1000_p4_points_distribution}). The duration between win milestones grows substantially: Table \ref{tab:1000_p4_player_milestones} shows Player 1 requires 123 games to reach 20 wins but 276 games to advance from 60 to 80 wins, with Player 2 showing similar patterns.  We have selected milestone in the context of total game scores, there is not reason to suggest that alternative segmentation produces different result. The final outcomes demonstrate significant deviation from optimal strategy. While uniform distribution, \(\frac{1}{3}\) probability, would yield approximately 333 points over 1000 games, Players 1 and 2 achieve only 92 and 104 points respectively. The 100-game averages of 25.3 and 26.5 points indicate a 10x increase in game length yields only a 3-4x increase in wins.
Rather than indicating a failure to learn, this behaviour may suggests deliberate loss-aversion strategies. Agents appear to pursue a minimax approach that priorities preventing opponent wins over maximising their own points, even when provided with complete game history. This is also evidenced in the lower than expected points of players in longer repeated games.  This interpretation aligns with established game theory concepts where players converge on loss minimising strategies in repeated interactions. 
Future research will explore whether this tendency toward defensive play is inherent to LLM design (possibly from human text parroting \cite{GebruParrot} or can be modified through inter-model game play.

\section{Prisoners Dilemma}
Next, we consider The Prisoner's Dilemma game (PD) as one-shot and repeated variants.  Unlike RPS, we are able to use the explicit payoff for co-operation to test LLM learning behaviours and compare this to two fundamental theoretical results: a unique dominant Nash equilibrium (Defect, Defect) and a Pareto optimal strategy of (Cooperate, Cooperate). Formally the game, \(G(PD)\), is defined as let \(G = (N, A, u)\) be a symmetric two-player normal-form game where: \(N = {1,2}\) is the set of players, \(A = {C,D}\) is the action set for each player, representing Cooperate and Defect respectively, and \(u: A \times A \rightarrow \mathbb{R}^2\) is the payoff function which can be expressed as:

\(\begin{array}{c|cc}
 & C & D  \\
\hline
C & (3,3) & (0,10) \\
D & (10,0) & (1,1) 

\end{array}  \)

Unlike RPS, this game has a unique pure strategy Nash equilibrium where both players choose to defect, despite mutual cooperation yielding a higher payoff. The dominant strategy equilibrium \(D,D)\) with payoff \((1,1)\) is strictly inferior to the mutual cooperation outcome \((C,C)\) with payoff \((3,3)\), creating the central dilemma that characterises this class of games.  We implement the code in the same structure at RPS (see Figure \ref{fig:game-flow}).  Taking inspiration from results on prompts discussed in RPS, we use the design of Prompt 1 for RPS as a template for the PD game prompts.

    \subsubsection*{Prompt PD-1: PD Base case}
        "\textit{You are playing a Prisoner's Dilemma game. Rules: Choose one of: 'C' (Cooperate) or 'D' (Defect). Payoff conditions: Both cooperate: you each get 3 points, Both defect: you each get 1 point, If you defect while other cooperates: you get 10 points and they get 0, If you cooperate while other defects: you get 0 points and they get 10 points. If you are given game history, analyze it to identify patterns. What is your choice? Only respond with a single letter: 'C' or 'D', no explanation}"

    \subsubsection*{Prompt PD-2: Express Strategy}
        Text of Prompt PD-1: PD Base case with the additional sentence \textbf{The Nash equilibrium for the game is mutual defection.}

\subsection{Results}
PD-1 focused on a base case test that found almost universal selection of the Pareto optimal choice (C,C) in both one-shot and repeated games. With one-shot cooperation at 93\% and repeated game cooperation at 100\%, there was no evidence of players learning tit-for-tat strategies. Notably, the Nash equilibrium (D,D) was never selected, hinting that the LLM may be cooperating or rather, in the one-shot case, that it may be aware of other player's welfare.
To test this hypothesis, PD-2 included the explicit statement that "The Nash equilibrium for the game is mutual defection." The results showed that in one-shot games, while (D,D) remained rare at 2\%, defection by at least one player occurred 24\% of the time. The repeated game results were particularly telling - 99\% of games converged to the Nash equilibrium (D,D), and 100\% of plays involved at least one defection. This comparison demonstrates that across repeated games with PD-2, the LLM appears to effectively learn how to achieve the Nash equilibrium, though this "success" resulted in the lowest possible stable scores (99,100) compared to the Pareto optimal outcomes in PD-1 (300,300).  Across one-shot games, LLMs appear more concerned with loss- aversion (and hence preferring C,C outcomes). Given the repeated patterns observed, we opted not to conduct tests at the 1,000-repetition game level.

\begin{table}[ht]
    \centering

    \begin{tabular}{l|l|cc|cc|}
        \cline{3-6}
        & & \multicolumn{2}{c|}{\makecell{\textbf{One-Shot} \\ \textbf{Games}}} & \multicolumn{2}{c|}{\makecell{\textbf{100 Repeated} \\ \textbf{Games}}} \\
        & & \textbf{C} & \textbf{D} & \textbf{C} & \textbf{D} \\
        \hline
        \multirow{4}{*}{\textbf{PD-1}} 
        & C & 93\% & 6\% & 100\% & 0\% \\
        & D & 1\% & 0\% & 0\% & 0\% \\
        \cline{2-6}
        & Player 1 Sum & \multicolumn{2}{c|}{289} & \multicolumn{2}{c|}{300} \\
        & Player 2 Sum & \multicolumn{2}{c|}{339} & \multicolumn{2}{c|}{300} \\
        \hline
        \multirow{4}{*}{\textbf{PD-2}} 
        & C & 76\% & 10\% & 0\% & 1\% \\
        & D & 12\% & 2\% & 0\% & 99\% \\
        \cline{2-6}
        & Player 1 Sum & \multicolumn{2}{c|}{350} & \multicolumn{2}{c|}{99} \\
        & Player 2 Sum & \multicolumn{2}{c|}{330} & \multicolumn{2}{c|}{100} \\
        \hline
    \end{tabular}
    \caption{Strategy Distribution and Cumulative Scores (100 Simulations)}
    \label{tab:pd_strategy_comparison2}
\end{table}


\section{Conclusions and Future work}   

In this research we test LLMs' capabilities and limitations in game play.  We systematic test two games: Rock Paper Scissors and Prisoner's Dilemma.  We demonstrate that while LLMs can engage in game-theoretic scenarios, they exhibit significant limitations in both randomisation and strategic adaptation.  We analyse one-shot and repeated forms of both games.  A variety of opportunistic prompts were tested, with mixed impacts. 
In RPS testing, LLMs consistently failed to reliably achieve uniform distribution in move selection (one-shot play) for all plays, however in repeated games their fared better, though still not meeting the theoretical uniform distribution for all moves. In both cases a statistically significant higher than expected number of games resulted in a tie.  Increasingly, the longer repeated games went on these models exhibited increasingly loss-averse strategies, preferring a stalemate and sub-optimal point accumulation compared to theoretical maxima. 

The Prisoner's Dilemma experiments revealed that again LLMs deviate from the game dominant strategy (and Nash equilibrium) in one-shot testing, with only marginal improvement in strategy with refined prompts. However, in repeated games the results were quite different -  LLMs can shift dramatically between cooperative and competitive strategies based on prompt design, particularly when explicitly informed about Nash equilibrium concepts.  Repeated games with a parsimonious prompt followed a Pareto optimal strategy at the expense of optimising for individual strategic results and thereby exhibiting a degree of agent co-operation.  However  PD-2, a prompt suggesting the strict game dominance strategy within a prompt, resulted in 100\% dominant game play.  No tit-for-tat strategy play was observed in either scenario.  The progression from cooperative strategies in basic prompts to competitive behaviour under equilibrium-specific prompting demonstrates both the adaptability and limitations of current LLM architectures.  It seems likely that in games or scenarios where randomness is required, careful use and monitoring of LLMS is necessary. In repeated games LLMs demonstrate learning abilities, though the objective of such learning is not necessarily the dominant strategic equilibrium. Loss-average requires special attention as this may also illustrate a degree of stalemate preference.  These results raise concerns for the use of LLMs in multi-agent systems where sustained strategic interaction is needed, as game play appears to converge to stable but sub-optimal outcomes rather than maintain dynamic strategic choices or follow a prori distributions.  

We observed important technical considerations for multi-agent LLM systems, particularly regarding API caching and response consistency.  We employed three techniques to preserve the ideal independent sampling of the LLM in game simulations through APIs; API key refresh, addition of unique variables to each API call and addition of unique game history (for repeated games).  We highlight the technical challenges and need for careful system design when deploying LLMs in scenarios requiring frequent sampling. 

Our findings demonstrate that while LLMs can engage in game-theoretic scenarios, they exhibit significant limitations in both randomisation and strategic adaptation. In our future research we plan to explore fine tuning approaches to improve both randomisation and strategic reasoning capabilities while preserving LLMs' core strengths.


\bibliographystyle{named}
\bibliography{references}

\end{document}